\documentclass[a4paper, 10pt, fleqn]{article}

\usepackage[a4paper, margin=1in]{geometry} % Margins
\usepackage{graphicx} % For figures
\usepackage{fancyhdr} % For headers and footers
\usepackage{titlesec} % For section formatting
\usepackage{setspace} % For spacing
\usepackage{amsmath, amssymb} % For math
\usepackage{enumitem} % For custom lists
\usepackage{hyperref} % For hyperlinks
\usepackage{totpages}
\usepackage{caption} % Required for caption formatting
\usepackage{float}   % For figure placement
\usepackage{booktabs} % For tables
\usepackage{algorithm} % For algorithm environment
\usepackage{algpseudocode} % For pseudocode commands
\usepackage{graphicx} % 导言区引入
\usepackage{subfigure}
\usepackage[table]{xcolor}
\usepackage{multirow}
\begin{document}
	\newcommand{\customauthor}[3]{\textbf{#1}\textsuperscript{#2}\ifnum#3=1\textsuperscript{}\fi}
	
	% Command to define an affiliation.
	% #1: Affiliation number (used to match with \customauthor)
	% #2: Full affiliation text (department, institution, address, etc.)
	% Example: \customaffiliation{1}{Department of Physics, XYZ University, Some City, Some Country}
	% This defines the affiliation corresponding to "1" in \customauthor.
	\newcommand{\customaffiliation}[2]{\textsuperscript{#1}\textit{#2}}
	
	% Author Details
	% This command lists all authors. Use the \customauthor command for each author.
	% Update the author names, their affiliation numbers, and corresponding author status.
	% Each author's details should be separated by a comma (,) in this command.
	\newcommand{\authorslist}{
		\customauthor{Liwen Zhang}{1}{1}, % Eva is the corresponding author, affiliated with #1
		\customauthor{Dong Zhou}{1}{1}, % Rock is a non-corresponding author, affiliated with #2
		\customauthor{Shibo Shao}{1}{1},           % Replace XXXXXX with another author's name, affiliated with #3
            \customauthor{Zihao Su}{1}{1},
            \customauthor{Guanghui Sun*}{1}{1}
	}
	
	% Affiliation Details
	% This command lists all affiliations. Use the \customaffiliation command for each unique affiliation.
	% Ensure that the affiliation number matches the one used in \customauthor.
	\newcommand{\affiliationslist}{
		\customaffiliation{1}{School of Astronautics, Harbin Institute of Technology, Harbin, China}
	}
\begin{center}
		% Replace the placeholder below with the official paper number provided by the organizing committee.
		%\textbf{Paper Number: 25}\\[1.5em]
		
		% Replace the title placeholder below with your manuscript title.
		% The title should be concise and less than 20 words.
		\textbf{\large Multimodal Spiking Neural Network for Space Robotic Manipulation}\\[1.5em]
		
		% % The \authorslist command automatically displays all the authors in the specified format.
		% % You only need to edit the \authorslist command above to update the names.
		 \authorslist\\[0em]
		
		% % The \affiliationslist command automatically displays all affiliations below the author names.
		% % You only need to edit the \affiliationslist command above to update the affiliations.
	   \affiliationslist\\[0em]
		
		% % This line automatically adds a note to indicate the corresponding author.
		% % Ensure the correct corresponding author is marked in the \authorslist command above.
		%\textit{* Corresponding Author}\\[2em]
\end{center}

	% Abstract Section
\renewenvironment{abstract}{
		\begin{center}
			\textbf{Abstract}
		\end{center}
		\noindent\begin{spacing}{1}
		}{
		\end{spacing}\vspace{1em}
	}
\begin{abstract}
This paper presents a multimodal control framework based on spiking neural networks (SNNs) for robotic arms aboard space stations. It is designed to cope with the constraints of limited onboard resources while enabling autonomous manipulation and material transfer in space operations. By combining geometric states with tactile and semantic information, the framework strengthens environmental awareness and contributes to more robust control strategies. To guide the learning process progressively, a dual-channel, three-stage curriculum reinforcement learning (CRL) scheme is further integrated into the system. The framework was tested across a range of tasks including target approach, object grasping, and stable lifting with wall-mounted robotic arms, demonstrating reliable performance throughout. Experimental evaluations demonstrate that the proposed method consistently outperforms baseline approaches in both task success rate and energy efficiency. These findings highlight its suitability for real-world aerospace applications.
\end{abstract}
	
	% Keywords

	% Nomenclature Section
	\section*{Nomenclature}
	\vspace{-.4em} % Reduce vertical space
    	\begin{flushleft}
		\begin{tabbing}
			% Format: symbol = description
			\hspace{1cm} \= \hspace{1cm} \= \kill % Tab stops
			$d$ \> = \> the Euclidean distances from the gripper center to the target object \\
			$d_{\text{lf}}, d_{\text{rf}}$ \> = \> the Euclidean distances from the left fingertip and right fingertip to the target object \\
			$\mathbf{p}_{\text{mid}}^{xz}, \mathbf{p}_{\text{obj}}^{xz}$ \> ~~~~= \> the projected position of the gripper center and the target object onto the X–Z plane\\
			$ y_{\text{lf}}, y_{\text{rf}}$ \> = \> the Y-coordinates of the left and right fingertips \\
			$y_{\text{mid}}, y_{\text{obj}}$ \> ~~ = \> the Y axis positions of the gripper center and the target object \\
                %$y_{\text{offset}}$ \> = \> the vertical displacement between the gripper and the target object along the Y-axis \\
			$\delta_i, \lambda_i, \alpha_i, \gamma_i$ \> ~~~~~ = \> the tunable parameters \\
                $\epsilon$ \> = \> a safety margin accounting for tolerable deviation in finger spacings \\
                $\text{gap}_{\text{eef}}$ \> = \> the Euclidean distance between the two fingertips \\
                $s_{\text{obj}}$ \> = \> the side length of the target object \\
                $\mathbf{f}_{\text{lf}}, \mathbf{f}_{\text{rf}}$ \> = \> the sensed contact forces at the left and right fingertips \\
                $\mathbf{d}_{\text{lf} \rightarrow \text{obj}}, \mathbf{d}_{\text{rf} \rightarrow \text{obj}}$ \> ~~~~~~~~~~~=\>  ~~~~ the normalized pointing direction from each fingertip to the target object \\
                $\mathbf{1}_{\text{lf\_t}}, \mathbf{1}_{\text{rf\_t}}$ \> ~~~=\>  binary signals indicating fingertip contact (1 if in contact, 0 otherwise) \\
                $\mathbf{y}_{\text{eef}}$\> ~~ = \> the end-effector's Y axis \\ 
                $\mathbf{y}$ \> = \> the world Y axis \\
                $h_0, h_1, h_2$ \>~~~ = \> $h_0$ is object height, $h_1$, $h_2$ are lift thresholds \\
                %$\beta_1$ \> = \> the threshold for the distance between the gripper center and the target object \\
                %$\beta_2$ \> = \> the maximum allowable distance between each fingertip and the object \\
                %$g$ \> = \> denotes the gripper aperture width, $g_{\text{max}} = s_{\text{obj}} + \epsilon$ \\
                %$T_{\text{lf}}, T_{\text{rf}}$ \> = \> the binary tactile contact signals, triggered based on force sensing thresholds \\
                $s_i^{(b)}$ \> = \> whether neuron $i$ fired in sample $b$ \\
                $V_i(t)$ \> = \> the membrane potential of neuron $i$ at time $t$ \\
                $\mathbb{I}(\cdot)$ \> = \> the indicator function \\
                $B$ \> = \> the batch size \\
                $T$ \> = \> the simulation steps \\
                $N_0, N_1, N_2$ \> ~~~~ = \> the layer sizes \\
                $x_i^{(b)}, h_j^{(b)}$ \> ~ = \> the activations of input and output neurons \\
                $\mathbf{f}_i$ \> = \> the 3D force vector measured at fingertip $i$ \\
                $f_{ix}, f_{iy}, f_{iz}$ \> ~~~~ = \> the components of the contact force along the X, Y, and Z coordinate axes \\
                $\mathbf{p}_i$ \> = \> the fingertip position \\  
                $\mathbf{c}$  \> = \> the object's center of mass
		\end{tabbing}
	\end{flushleft}
	% Acronyms/Abbreviations Section
	% \section*{Acronyms/Abbreviations}
	% This section is not numbered. Define acronyms and abbreviations that are not standard in this section. Such acronyms and abbreviations that are unavoidable in the abstract must be defined at their first mention there. Ensure consistency of abbreviations throughout the article. Always use the full title followed by the acronym (abbreviation) to be used, e.g., reusable suborbital launch vehicle (RSLV), International Space Station (ISS). List acronyms and abbreviations in alphabetical order.
	
	% Introduction Section
\section{Introduction}

As crewed spaceflight, Mars exploration, and on-orbit servicing advance, space manipulators have become essential to spacecraft operations. They are generally categorized as extravehicular or intravehicular~\cite{yang2021manned}. Extravehicular manipulators, with extended reach and high robustness, serve in tasks such as assembly, maintenance, and target capture~\cite{dou2023disturbance, virgili2019simultaneous, gangapersaud2019detumbling}. Intravehicular manipulators are adapted to confined environments and enable fine manipulation within spacecraft cabins~\cite{yamaguchi2024design, zhang2023omnidirectional}.

Conventional control approaches rely on accurate modeling and vision-based guidance~\cite{aghili2022automated, aghili2012active, faust2013learning}. Recently, artificial neural networks (ANNs) with reinforcement learning (RL) has attracted increasing attention for space manipulator control~\cite{jahanshahi2023unified, cao2023reinforcement, hu2025deep, zhou2023deep, zhou2022space}. However, ANN-based methods often involve high computational cost and latency, which limit their suitability for space missions requiring real-time performance and energy efficiency.

Spiking neural networks (SNNs) have demonstrated promising energy efficiency in terrestrial robotics. However, their application in space manipulator control remains an open research question. Motivated by this gap, we develop a control framework that integrates SNNs with RL. The framework examines the feasibility of using SNNs for energy-efficient control, multimodal perception, and task execution under typical space-related constraints.

The structure of this paper is as follows: Section~\ref{sec:level2} introduces the control approach. Section~\ref{sec:level3} presents the experiments and results. Section~\ref{sec:level4} concludes the work and discusses future directions.

% Material and Methods
\section{\label{sec:level2}Methodology}
\subsection{Model Architecture}

\begin{figure}[htbp]
    \centering
    \includegraphics[width=1\linewidth]{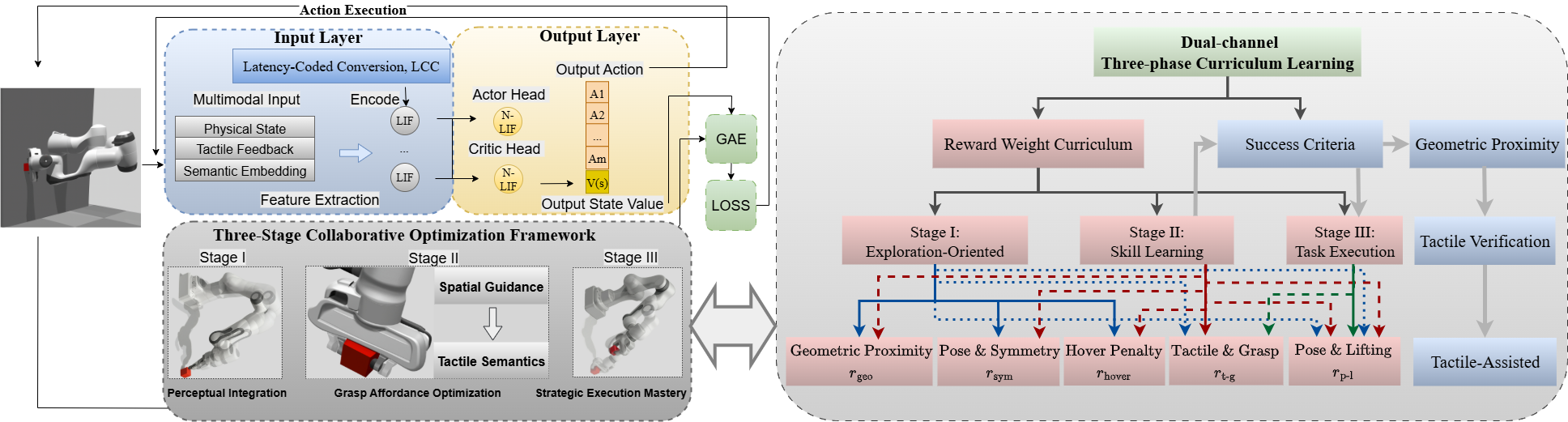}
    \caption{The system architecture.}\label{fig1}
\end{figure}

A control architecture for robotic manipulators in space station environments is designed in this study, inspired by the SpikeGym framework~\cite{zanatta2024exploring}. A schematic of the overall system is shown in the left part of Fig.~\ref{fig1}.

The proposed control framework incorporates three categories of sensory input: physical states, tactile signals, and semantic cues. Environmental observations are encoded into spike trains by leaky integrate-and-fire (LIF) neurons at the input layer. These spike patterns are then processed by a non-spiking LIF (N-LIF) output layer, which generates signals for action selection and value estimation.

The control process is structured in three stages. In the first stage, spatial features are integrated to guide accurate target approach. The second stage incorporates tactile force information and high-level semantic information, enabling the system to model and infer grasp affordance more effectively. In the third stage, the framework shifts from purely geometric representations to a multimodal learning mechanism, allowing the robot to adapt more robustly to dynamic and uncertain interactions. Finally, the system is trained in an end-to-end manner using proximal policy optimization (PPO) with generalized advantage estimation (GAE). 

% \begin{figure}[htbp]
%     \centering
%     \includegraphics[width=0.4\linewidth]{framework.drawio.png}
%     \caption{Overview of the System Architecture.}\label{fig1}
% \end{figure}

\subsection{Tactile Feature Representation}
This work proposes a set of tactile features to characterize contact interactions during robotic grasping, motivated by the insights discussed in the review by \cite{jahanshahi2025comprehensive}. Specifically, we define:
(1) the norm of the contact force at each fingertip is used to quantify local contact intensity;
(2) the symmetry of force magnitudes between the two fingertips indicates force balance;
and (3) the geometric consistency is evaluated via the directional alignment between the applied contact force $\mathbf{f}_i$ and the positional vector from the fingertip location $\mathbf{p}_i$ to the object’s center of mass $\mathbf{c}$. The formal definitions of these features are provided below.

\setlength{\abovedisplayskip}{1pt}
\setlength{\belowdisplayskip}{1pt}
\begin{equation}
\|\mathbf{f}_i\| = \sqrt{f_{ix}^2 + f_{iy}^2 + f_{iz}^2},~s = \left| \|\mathbf{f}_{\text{lf}}\| - \|\mathbf{f}_{\text{rf}}\| \right|,~\cos(\theta_i) = \frac{\mathbf{f}_i}{\|\mathbf{f}_i\|} \cdot \frac{\mathbf{c} - \mathbf{p}_i}{\|\mathbf{c} - \mathbf{p}_i\|}.
\end{equation}

\subsection{Training Method}

To address the instability induced by unexpected drift and rotation of target objects during contact in microgravity, we introduce a dual-channel, three-phase curriculum reinforcement learning (CRL) framework (see the right part of Fig.~\ref{fig1}). The framework guides policy learning through staged modulation of reward structure and success criteria during training. The early stage emphasizes geometric alignment. As training progresses, the focus shifts toward tactile-guided control in the intermediate phase. In the final stage, the reward design centers on task completion as the primary objective. This structured curriculum enhances learning efficiency and improves policy robustness under space-relevant dynamic uncertainties.

% \begin{figure}[htbp]
% \centering
% \includegraphics[width=0.5\linewidth]{reward_neural_drawio.drawio.png}
% \caption{Overview of the dual-channel, three-stage curriculum framework.}\label{fig2}
% \end{figure}

The reward structure consists of five components: geometric alignment, tactile feedback and grasp execution, lift execution under posture constraints, and penalty terms.

\paragraph{(I) Exploration-Oriented} 

The main reward terms are defined as follows. The geometric proximity reward ($r_{\text{geo}}$) encourages the end-effector to maintain spatial closeness to the target. The pose and symmetry regularization term ($r_{\text{sym}}$) promotes alignment accuracy. Additionally, the failure aversion penalty ($r_{\text{hover}}$) penalizes actions that result in ineffective or unstable behavior.

\setlength{\abovedisplayskip}{1pt}
\setlength{\belowdisplayskip}{1pt}
\begin{equation}
r_{\text{geo}} = \sum_{i=1}^{3} \alpha_i \left(1 - \tanh(\lambda_i \cdot d_i)\right), d_1 = \frac{d + d_{\text{lf}} + d_{\text{rf}}}{3}, d_2 = \left\| \mathbf{p}_{\text{mid}} - \mathbf{p}_{\text{obj}} \right\|, d_3 = \left\| \mathbf{p}_{\text{mid}}^{xz} - \mathbf{p}_{\text{obj}}^{xz} \right\|.
\end{equation}

\setlength{\abovedisplayskip}{1pt}
\setlength{\belowdisplayskip}{1pt}
\begin{equation}
r_{\text{sym}} = \alpha_4 \left(1 - \tanh \left( \lambda_4 \cdot |y_{\text{lf}} - y_{\text{rf}}| \right) \right) 
+ \gamma_1 \cdot \max\left(0, \delta_1 - |y_{\text{mid}} - y_{\text{obj}}| \right).
\end{equation}

\setlength{\abovedisplayskip}{1pt}
\setlength{\belowdisplayskip}{1pt}
\begin{equation}
r_{\text{hover}} = - \exp\left( \lambda_5 \cdot \left( |y_{\text{mid}} - y_{\text{obj}}| - \delta_2 \right) \right).
\end{equation}

\paragraph{(II) Skill Learning} 
The tactile-grasp consistency reward ($r_{\text{t-g}}$) encourages balanced contact and stable grasping of the object. The condition for grasp success, transitioning progressively from a geometry-based assessment to one informed by tactile cues. 
\setlength{\abovedisplayskip}{1pt}
\setlength{\belowdisplayskip}{1pt}

\begin{align}
r_{\text{t-g}}\!\!= \!\!\alpha_5 \exp(-\lambda_6\left| \text{gap}_{\text{eef}}\!\! -\!\! (s_{\text{obj}} \!\!+\!\! \epsilon) \right|) \!\!+ \!\!\frac{1}{2}\big[(1 \!\!- \!\!\tanh(\lambda_7\left| \left\| \mathbf{f}_{\text{lf}} \right\| \!\!-\!\! \left\| \mathbf{f}_{\text{rf}} \right\| \right|))\!\!+\!\!( \mathbf{f}_{\text{lf}}\mathbf{d}_{\text{lf} \rightarrow \text{obj}} \!\!+\!\! \mathbf{f}_{\text{rf}}  \mathbf{d}_{\text{rf} \rightarrow \text{obj}})\!\!+\!\!(\mathbf{1}_{\text{lf\_t}} \!\!+\!\!\mathbf{1}_{\text{rf\_t})}\big]
\end{align}

\paragraph{(III) Task Execution} 

The primary objective at this stage is to lift the object while maintaining grasp stability, which has already been ensured through tactile feedback. To achieve this, a pose-lift guidance reward ($r_{\text{p-l}}$) is introduced. This reward encourages proper orientation of the end-effector, and facilitates a continuous and stable lifting motion throughout the manipulation process.

\begin{equation}
r_{\text{p-l}} = \alpha_8 \cdot \mathbf{y}_{\text{eef}} \cdot (-\mathbf{y}) + \alpha_9 \cdot \min \left( \gamma_2 \cdot (h + \delta_3), \gamma_2 \right).
\end{equation}

\subsection{Estimation of Energy Consumption} \label{energy}

This section introduces an analytical model designed to compare the computational energy consumption of SNNs and ANNs. The approach is grounded in operation-level energy profiling, as detailed in previous studies~\cite{rathi2021diet, rueckauer2017conversion}. In SNNs, the encoder layer typically performs floating-point multiply-accumulate (MAC) operations. In contrast, the following fully connected layers mainly rely on accumulate (AC) operations. ANNs, on the other hand, employ MAC operations uniformly across all layers~\cite{jiang2025fully}.

The estimated total energy consumption of SNN is:
\begin{equation}
   E_{\text{SNN}} = B T \big[ N_1 (\frac{1}{B N_1} \sum_{b=1}^{B} \sum_{i=1}^{N_1} s_i^{(b)}) (N_0 \alpha_m + (N_0{-}1) \alpha_a) + N_2 (\frac{1}{N_2} \sum_{i=1}^{N_2} \mathbb{I}\left(\sum_{t=1}^{T}|V_i(t)|>0\right))) N_1 \alpha_a \big] 
\end{equation}
where \(\alpha_m\), \(\alpha_a\) are the energy costs of multiplication and addition (set to 4.6 and 0.9 pJ~\cite{kundu2021hire, yin2021accurate, yao2023attention, jiang2025fully}).

The estimated total energy consumption of ANN is:
\begin{align}\label{eq:ann_energy}
E_{\text{ANN}} \!\!=\!\! B T [ 
& N_1 (\frac{1}{B N_0} \sum_{b,i} \mathbb{I}(x_i^{(b)} \!\!>\!\! 0)) (N_0 \alpha_m \!\!+\!\! (N_0{-}1) \alpha_a) \!\!+\!\! N_2 (\frac{1}{B N_1} \sum_{b,j} \mathbb{I}(h_j^{(b)} \!\!>\!\! 0)) (N_1 \alpha_m \!\!+\!\! (N_1{-}1) \alpha_a) ]
\end{align}

\section{\label{sec:level3}Experiments and Discussion}

This section presents the performance evaluation of a wall-mounted manipulator designed for operation within a space station. The assessment focuses on its ability to perform target-reaching, grasping, and lifting tasks. Experiments were carried out in the Isaac Gym simulation platform \cite{makoviychuk2021isaac}, where 8,192 parallel instances were deployed to accelerate training and improve sample efficiency.

This study evaluates the performance of four control models: SNN and ANN architectures under both multimodal and unimodal input conditions. Each model was tested in 10 independent trials. In every trial, the target object's position was randomly initialized to ensure robustness and generalization of the learned policy.

Performance was assessed using two primary metrics: task success rate and policy stability. The success rate for grasping and lifting was defined as the proportion of parallel environments that successfully completed the task during a single evaluation. Policy stability was measured by examining the convergence behavior of the reward signal over the course of training.

\begin{figure}[htbp]
\centering
\subfigure[]{\includegraphics[width=0.25\linewidth]{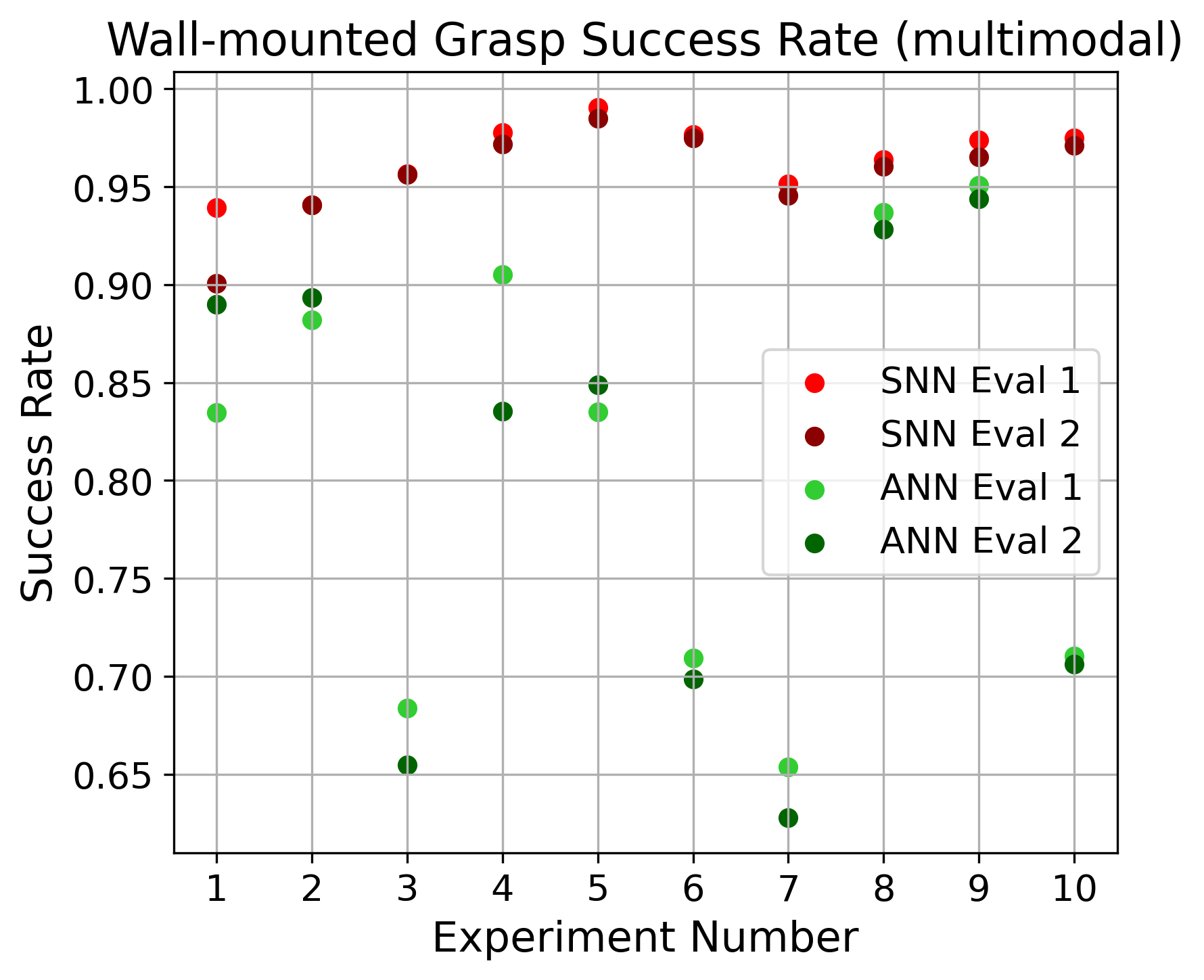}}
\subfigure[]{\includegraphics[width=0.245\linewidth]{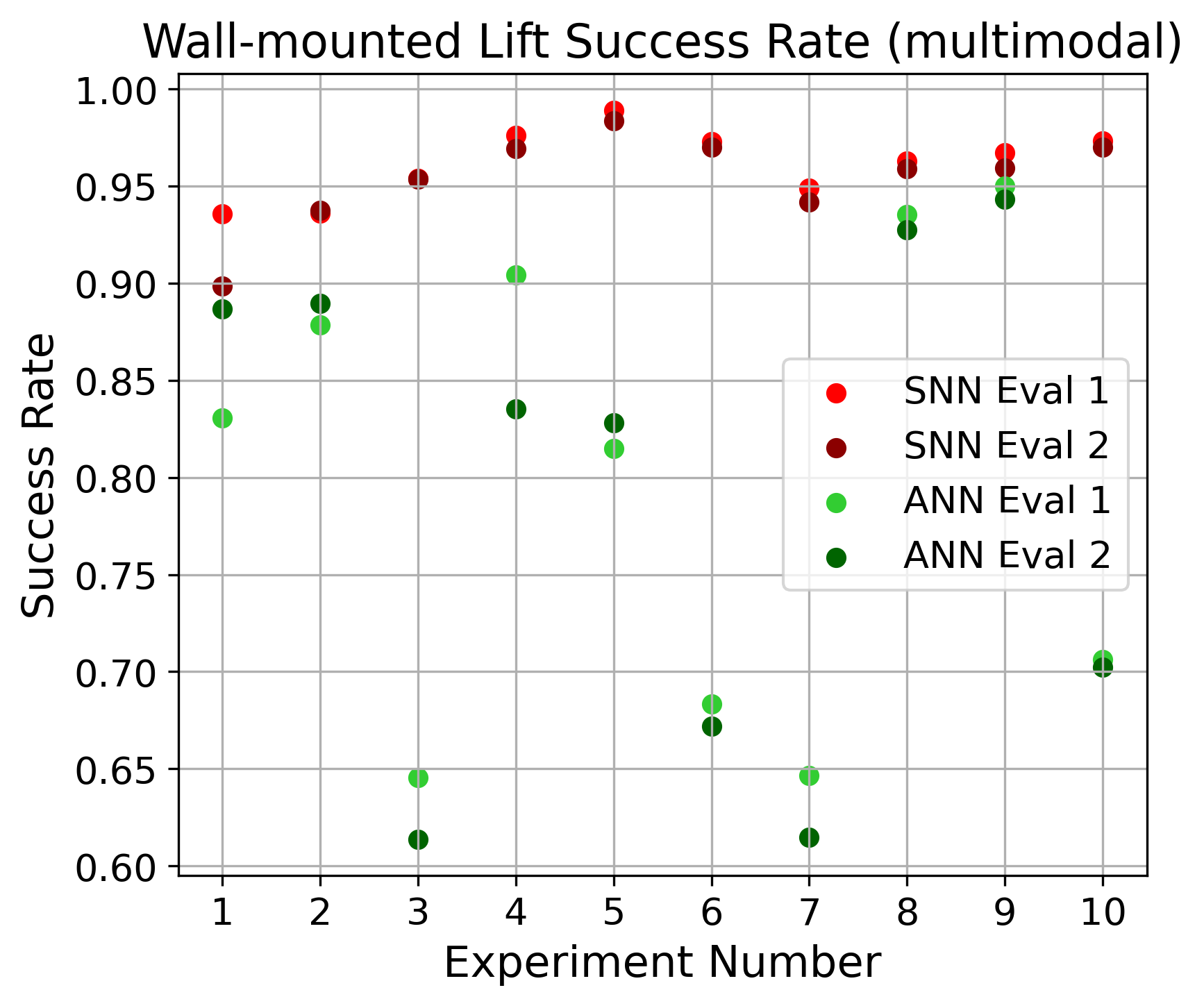}}
\subfigure[]{\includegraphics[width=0.24\linewidth]{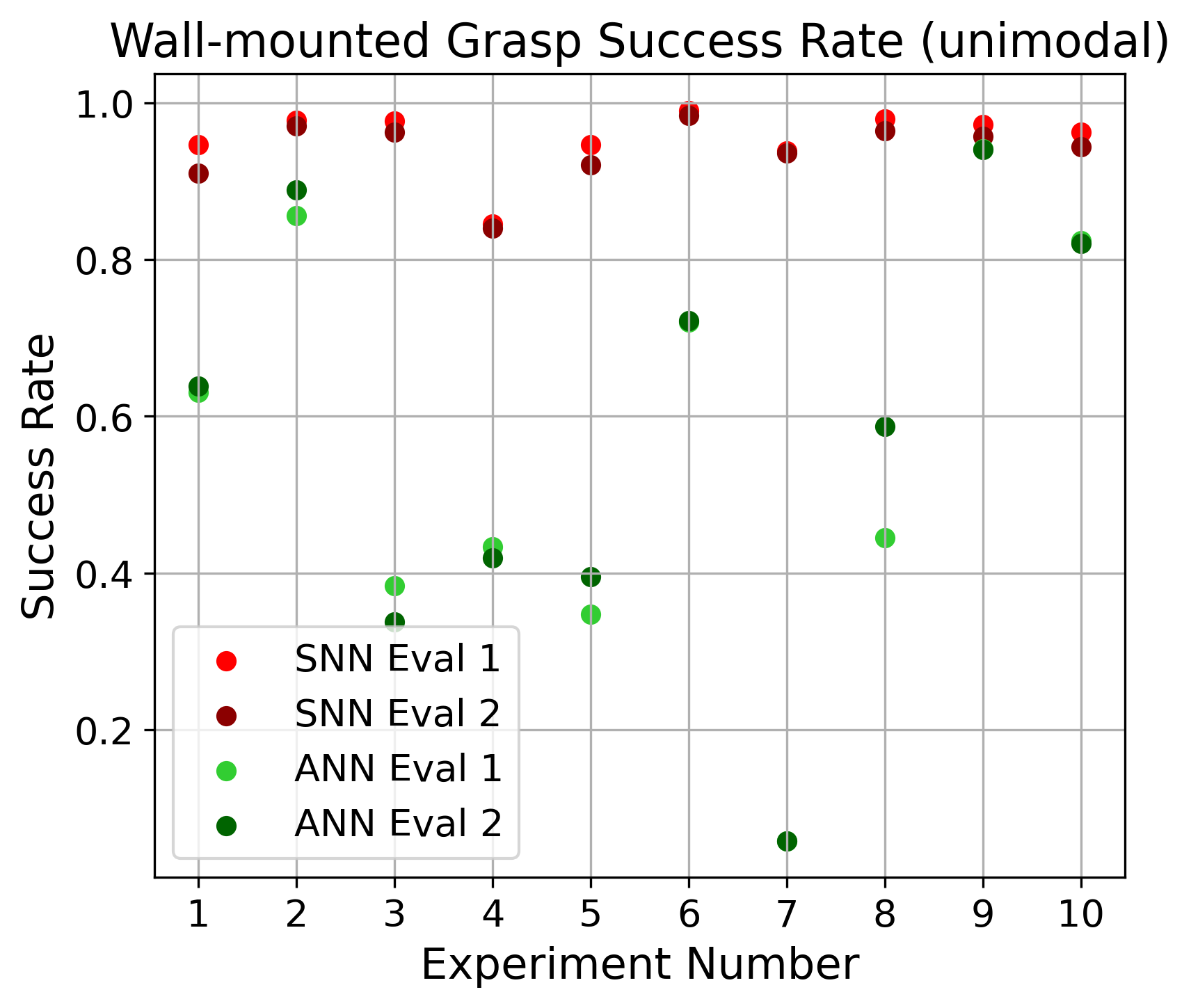}}
\subfigure[]{\includegraphics[width=0.235\linewidth]{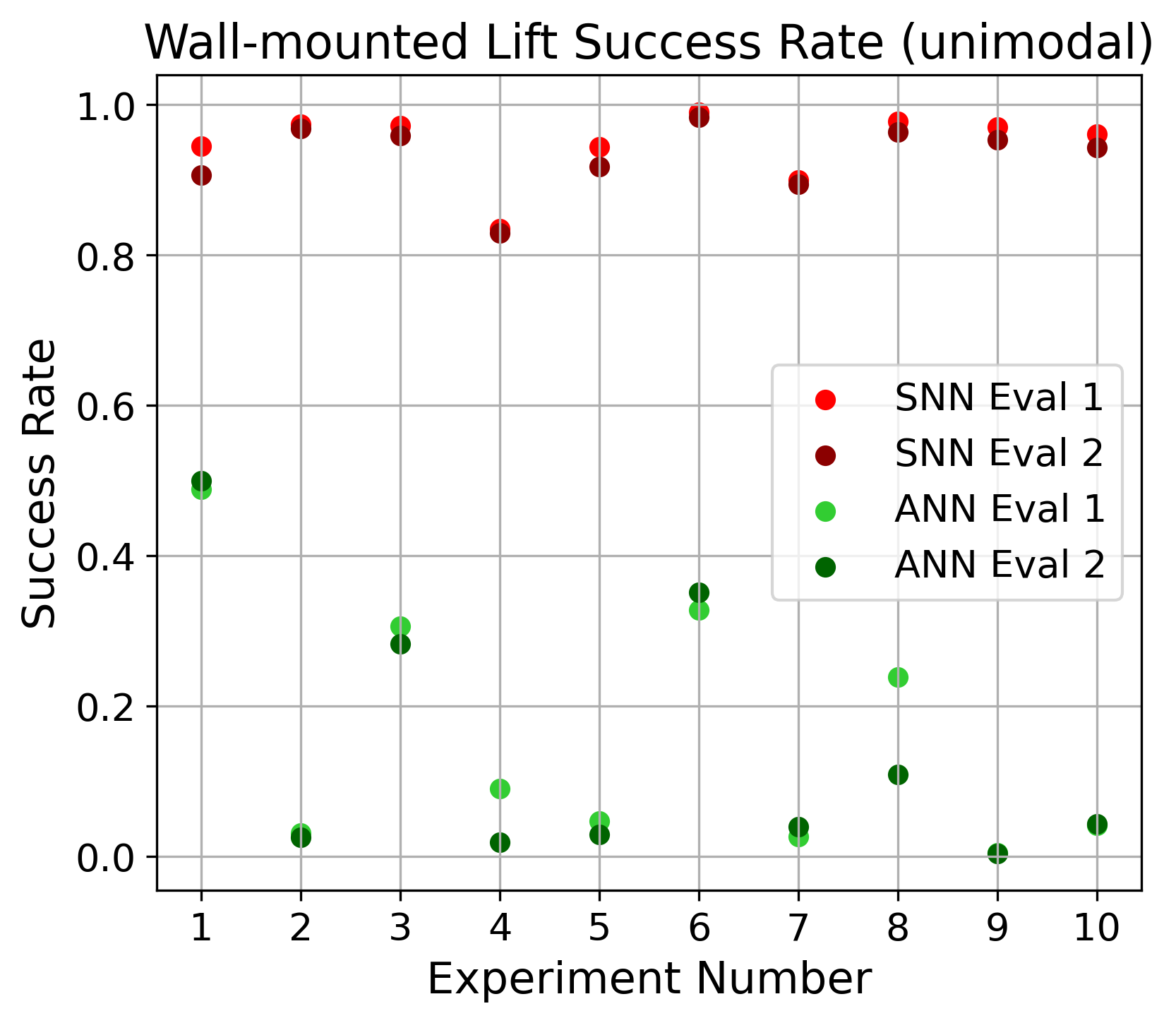}}
\caption{In multimodal and unimodal scenarios, SNN and ANN (each with two evaluations) were tested in 10 experiments for the success rates of grasping and lifting tasks.}
\label{fig:4subfigs}
\end{figure}

\begin{figure}[htbp]
\centering
\subfigure[]{\includegraphics[width=0.23\linewidth]{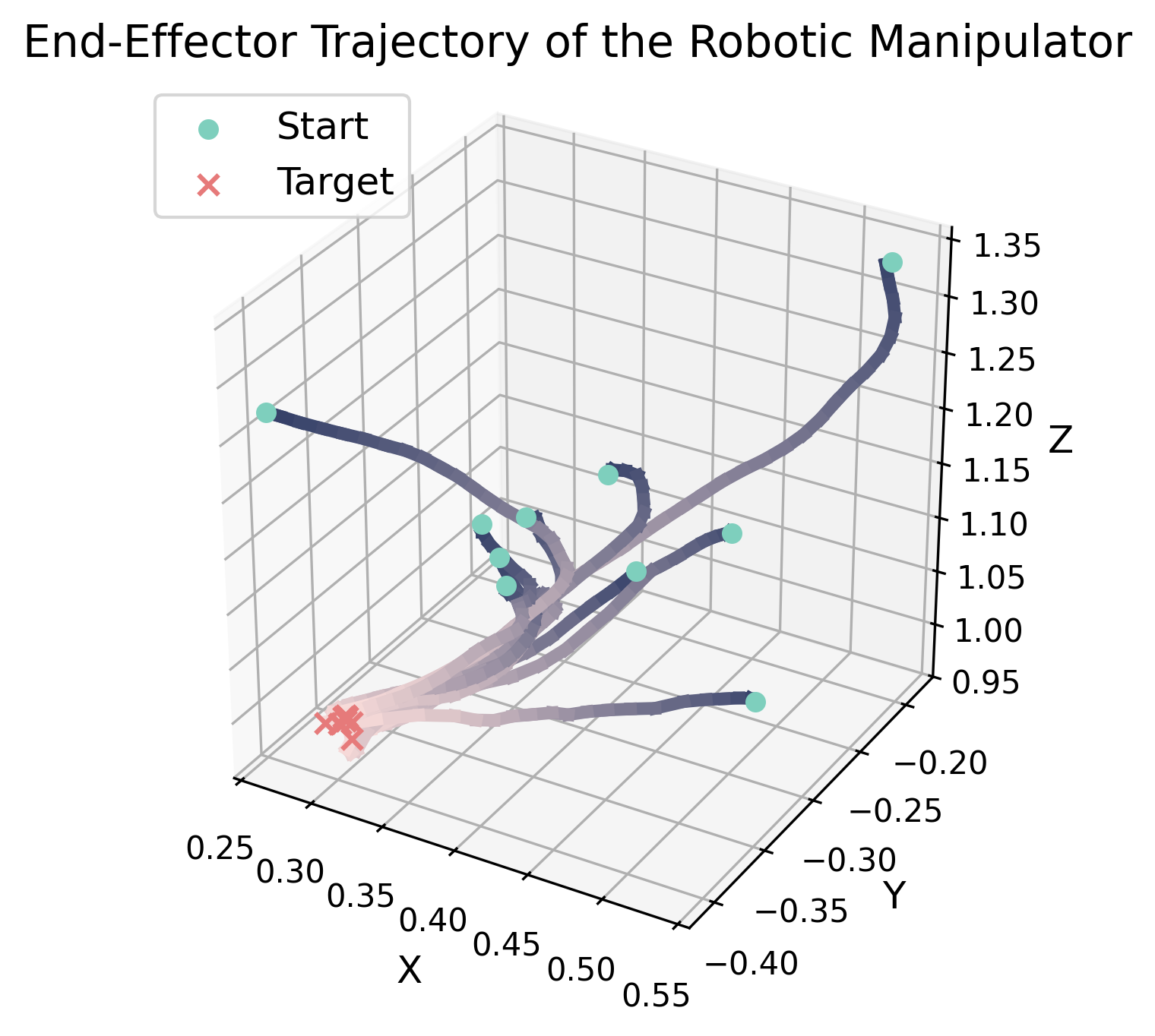}}
\subfigure[]{\includegraphics[width=0.335\linewidth]{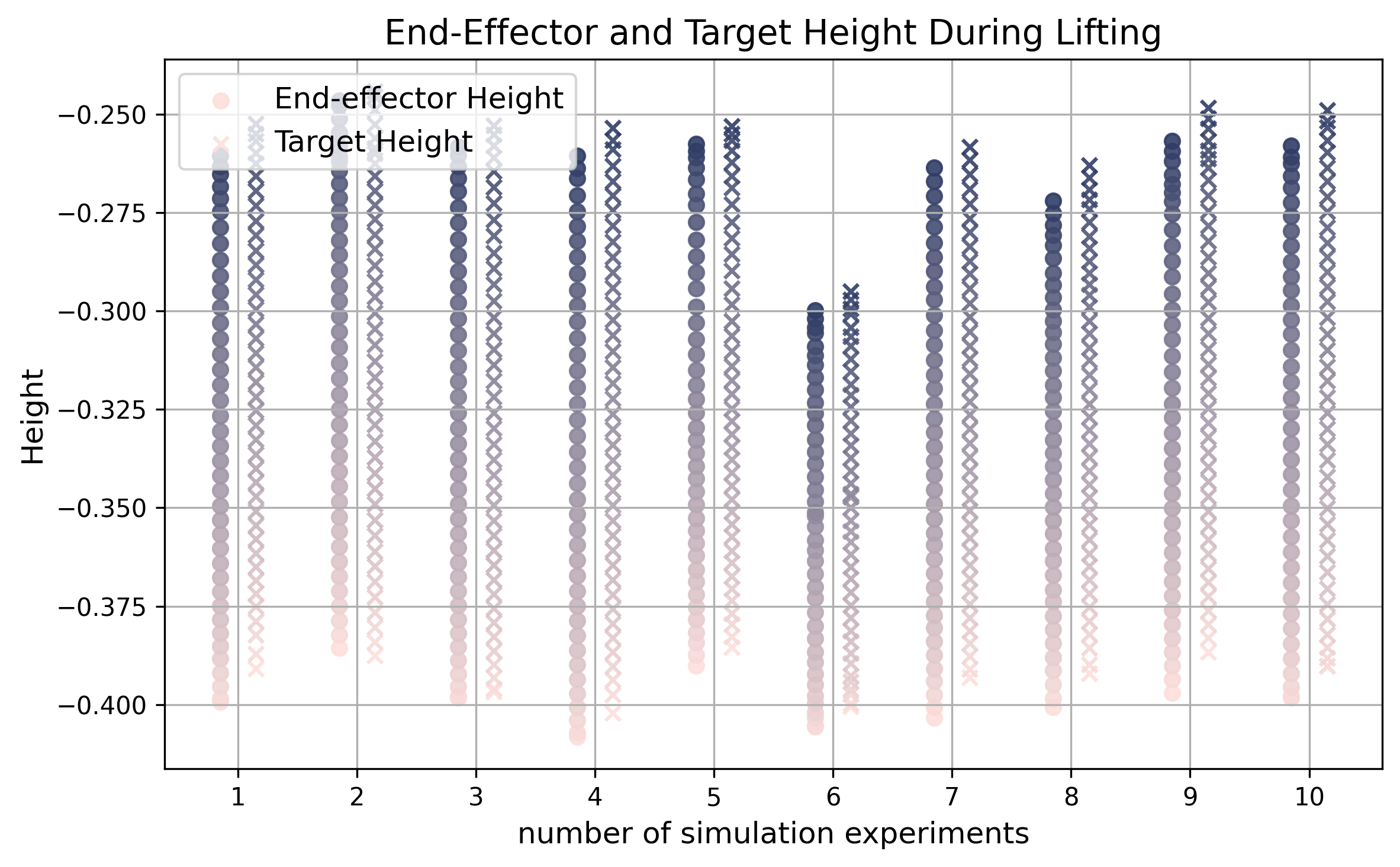}}
\caption{Trajectory schematic between end effector and target during space manipulator grasping and lifting tasks.}
\label{fig:2subfigs}
\end{figure}

As illustrated in Fig.~\ref{fig:4subfigs}, the multimodal SNN control framework demonstrates overwhelming advantages in space manipulator operation tasks. In the grasping task under multimodal input conditions (Fig.~\ref{fig:4subfigs}(a)), the grasping success rate of the SNN model is significantly higher than that of the ANN model under the same conditions. When using unimodal input (Fig.~\ref{fig:4subfigs}(c)), the success rate advantage of the SNN further expands, while the performance of the ANN fluctuates more significantly. The lifting task further corroborates similar conclusions (Fig.~\ref{fig:4subfigs}(b), (d)). Additionally, Fig~\ref{fig:2subfigs}(a) and Fig~\ref{fig:2subfigs}(b) present the motion trajectory of the end-effector approaching the target object in the grasping task, and the dynamic variation of the target height over time in the lifting task, respectively.

% In the lifting task, the multimodal SNN (Figure 2(b)) also shows notably superior lifting success rate, demonstrating an improvement of approximately 10\%-15\% compared to the ANN. Under unimodal scenarios (Figure 2(d)), the SNN maintains a stable advantage, whereas the success rate of the ANN drops below 0.5. This discrepancy highlights the robustness and adaptability of the multimodal SNN framework in complex robotic operation tasks, particularly in scenarios with limited or single-modal sensory inputs.

\begin{figure}[htbp]
\centering
\subfigure[]{\includegraphics[width=0.285\linewidth]{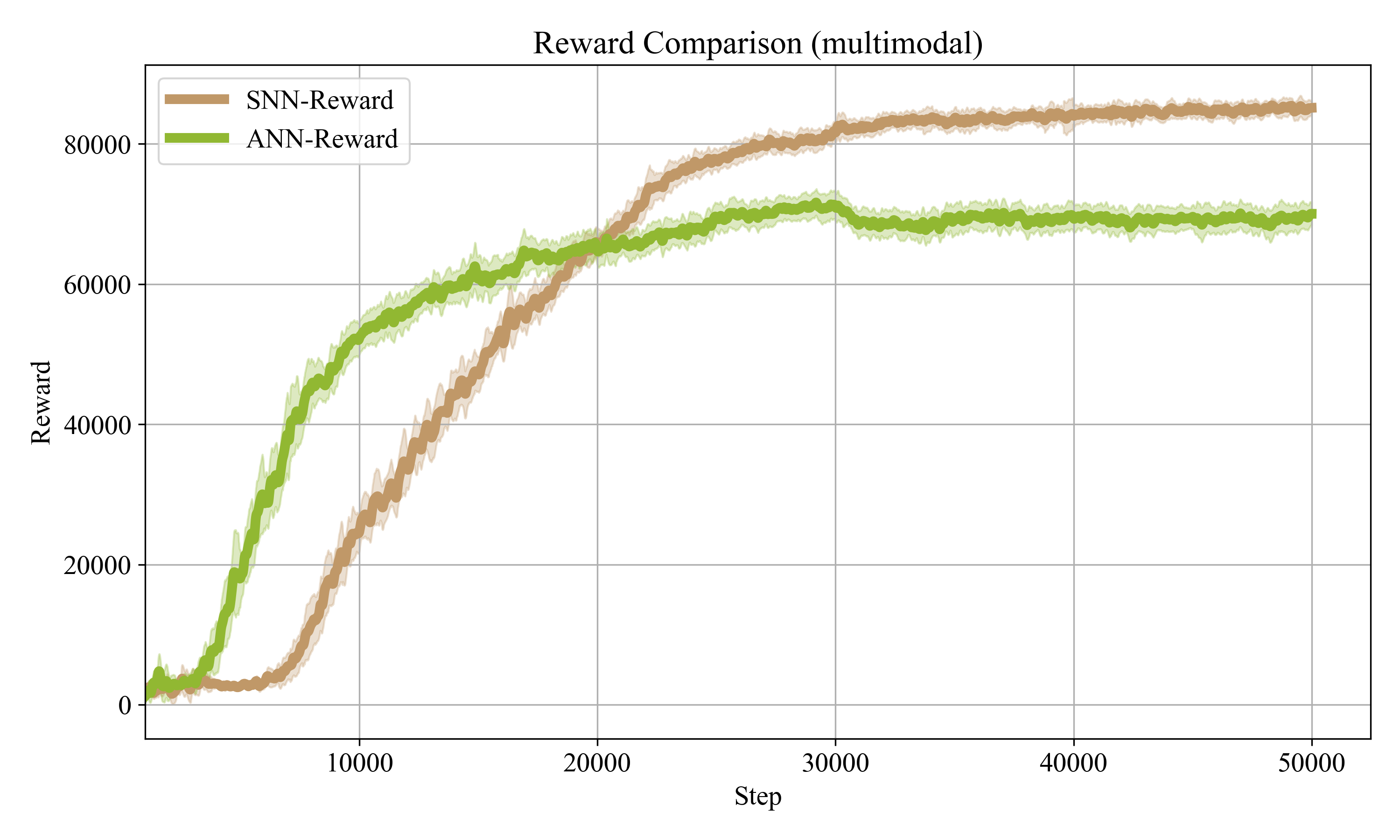}}
\subfigure[]{\includegraphics[width=0.285\linewidth]{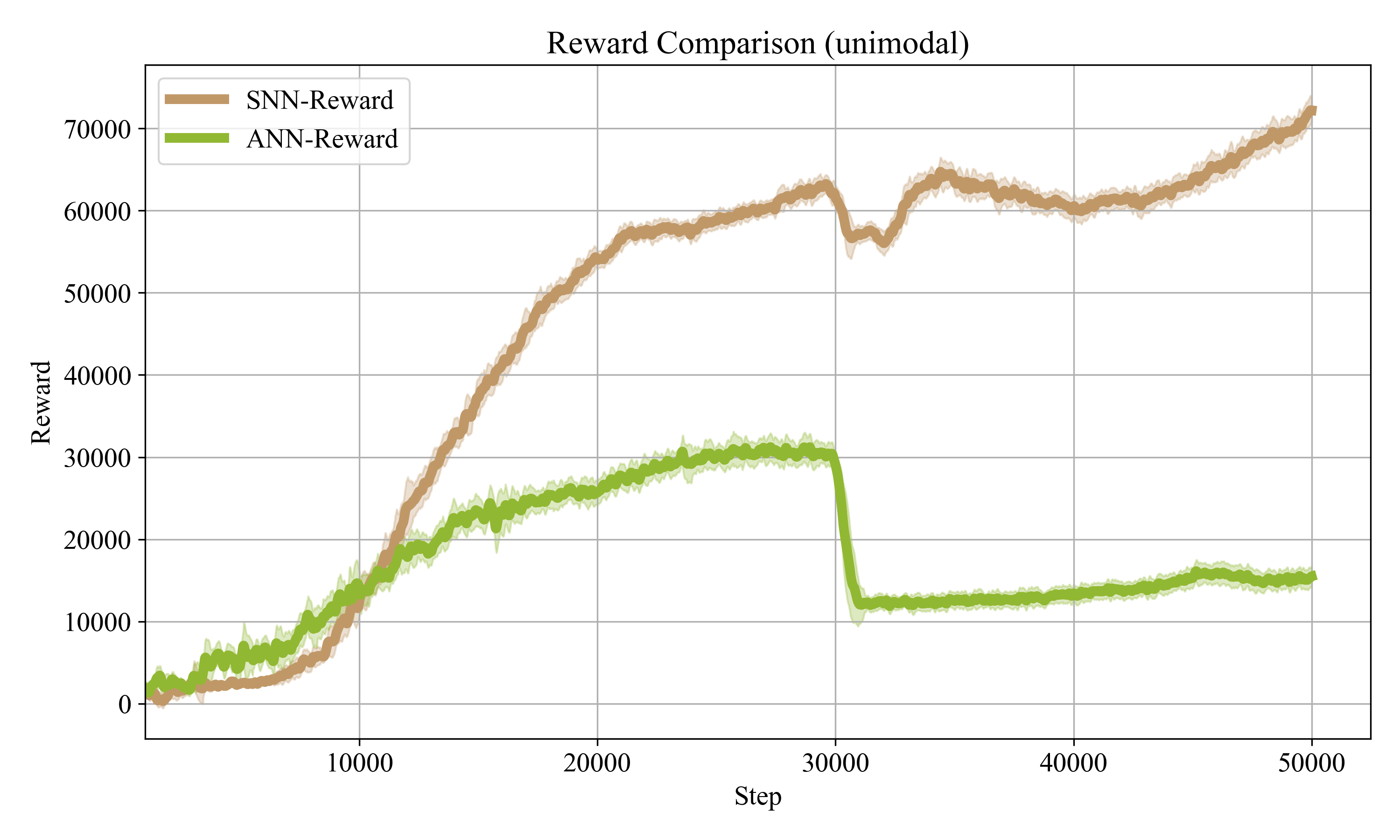}}
\subfigure[]{\includegraphics[width=0.366\linewidth]{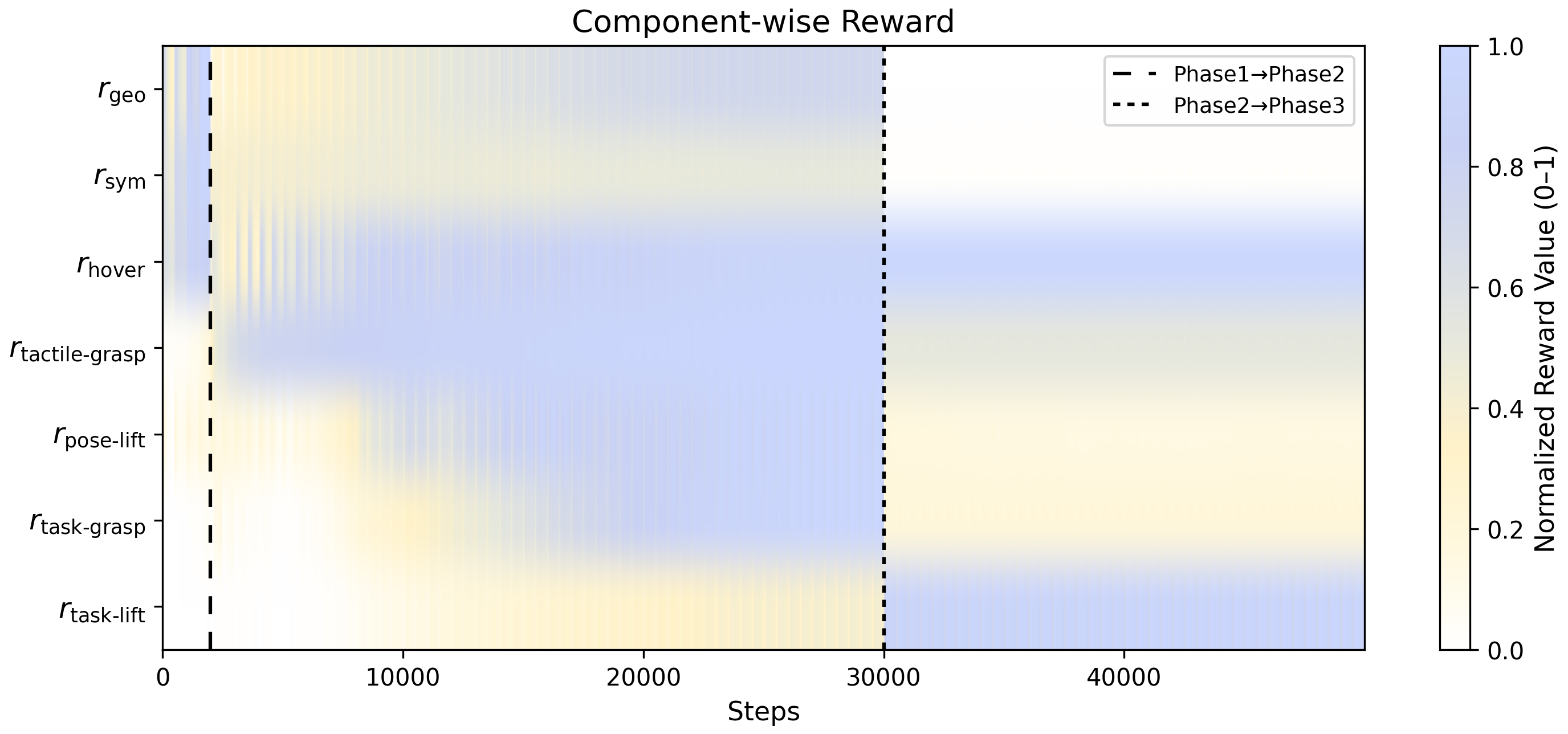}}
\caption{Overview of the reward variations.}
\label{fig:3subfigs}
\end{figure}

Besides the task success rate analysis, Fig.~\ref{fig:3subfigs}(a) and Fig.~\ref{fig:3subfigs}(b) present the reward variations, while Fig.~\ref{fig:3subfigs}(c) shows the dynamic evolution of each reward in the three-stage CRL of SNN under multimodal input. These data provide important evidence for exploring the learning efficiency and convergence behavior of strategies implemented by SNN and ANN. The results indicate that in multimodal input scenarios, both SNN and ANN demonstrate robust training stability, and the SNN learning strategy shows higher effectiveness. In contrast, under unimodal input conditions, the training stability of both SNN and ANN deteriorates significantly.

% \begin{figure}[htbp]
% \centering
% \includegraphics[width=0.5\linewidth]{reward_heatmap_custom_bluered0630_1808.png}
% \caption{Overview of the dual-channel, three-stage curriculum framework.}\label{fig2}
% \end{figure}

To evaluate the energy efficiency of the proposed SNN framework, we performed a comparative study against an ANN baseline under identical robotic manipulation tasks. The energy consumption was estimated using an analytical model, which accounts for both the type of arithmetic operations and the level of activation sparsity.
\begin{table}[htbp]
\centering
\small % 或者 \footnotesize, \scriptsize
\renewcommand{\arraystretch}{1} % 行间距可再调小如0.9
\resizebox{0.7\linewidth}{!}{ % 调整为表格不占满整行
\begin{tabular}{|c|c|c|c|c|c|c|c|c|}
\hline
\textbf{Model} & ${r(r_{in})}$ & ${r_{mem}(r_{\text{out})}}$ & $B$ & $T$ & $N_0$ & $N_1$ & $N_2$ & $E_{\text{final(mJ)}}$ \\
\hline
SNN & 0.34 & 1 & 8192 & 500 & 29 & 256 & 7 & 63149.44 \\
ANN & 1 & 0.44 & 8192 & 500 & 29 & 256 & 7 & 184055.68 \\
\hline
\multicolumn{8}{|l|}{\textbf{Energy Saving}} & \textbf{65.69\%} \\
\hline
\end{tabular}
}
\caption{Energy Consumption Comparison.}
\label{table energy}
\end{table}

\section{\label{sec:level4}Conclusion}
This paper proposes a fully spiking actor-critic framework enhanced by a two-channel, three-stage CRL strategy. The framework enables efficient control operations of robotic arms in space stations. It provides technical support to alleviate the in-cabin labor burden of astronauts.

Experimental evaluations demonstrate that SNN-based agents outperform their ANN counterparts in both training stability and overall task success rates. Additionally, the framework significantly boosts task success rates by integrating multimodal inputs based on tactile forces, thereby enhancing adaptation robustness to complex space environments.

Overall, this study proposes a robotic arm control method tailored for resource-constrained space environments. The approach offers a low-risk and efficient solution for autonomous operations, material transfer, and equipment-assisted tasks in scenarios such as space stations and planetary bases. Future research will focus on generalizing the framework to a wider set of manipulation tasks and exploring deployment on neuromorphic hardware platforms.
\bibliographystyle{elsarticle-num}  
\bibliography{snn}% Produces the bibliography via BibTeX.	
\end{document}